\documentclass[10pt,twocolumn,letterpaper]{article}
\usepackage{cvpr}              % To produce the CAMERA-READY version
\definecolor{cvprblue}{rgb}{0.21,0.49,0.74}
\usepackage[pagebackref,breaklinks,colorlinks,allcolors=cvprblue]{hyperref}

\hypersetup{
    pdftitle={FIESTA: Fisher Information-based Efficient Selective Test-time Adaptation},
    pdfauthor={Mohammadmahdi Honarmand, Onur Cezmi Mutlu, Parnian Azizian, Saimourya Surabhi, Dennis P. Wall},
    pdfkeywords={Facial Expression Recognition, Test-time Adaptation, Fisher Information, Domain Shift},
    pdfsubject={Computer Vision – Test-Time Adaptation in Affective Computing}
}

\def\paperID{*****} % *** Enter the Paper ID here
\def\confName{CVPR}
\def\confYear{2025}

\title{FIESTA: Fisher Information-based Efficient Selective Test-time Adaptation}

\author{
\begin{tabular}{c}
Mohammadmahdi Honarmand \quad Onur Cezmi Mutlu \quad Parnian Azizian \\
Saimourya Surabhi \quad Dennis P. Wall\thanks{Corresponding author} \\
Stanford University \\
{\tt\small \{mhonar, cezmi, azizian, mourya, dpwall\}@stanford.edu}
\end{tabular}
}

\usepackage{epstopdf}
\begin{document}
\maketitle
\begin{abstract}
Robust facial expression recognition in unconstrained, "in-the-wild" environments remains challenging due to significant domain shifts between training and testing distributions. Test-time adaptation (TTA) offers a promising solution by adapting pre-trained models during inference without requiring labeled test data. However, existing TTA approaches typically rely on manually selecting which parameters to update, potentially leading to suboptimal adaptation and high computational costs. This paper introduces a novel Fisher-driven selective adaptation framework that dynamically identifies and updates only the most critical model parameters based on their importance as quantified by Fisher information. By integrating this principled parameter selection approach with temporal consistency constraints, our method enables efficient and effective adaptation specifically tailored for video-based facial expression recognition. Experiments on the challenging AffWild2 benchmark demonstrate that our approach significantly outperforms existing TTA methods, achieving a 7.7\% improvement in F1 score over the base model while adapting only 22,000 parameters—more than 20 times fewer than comparable methods. Our ablation studies further reveal that parameter importance can be effectively estimated from minimal data, with sampling just 1-3 frames sufficient for substantial performance gains. The proposed approach not only enhances recognition accuracy but also dramatically reduces computational overhead, making test-time adaptation more practical for real-world affective computing applications.
\end{abstract}    
\section{Introduction}
\label{sec:intro}

Robust facial expression recognition in unconstrained, “in‐the‐wild” environments has become increasingly critical with the advent of applications such as driver-monitoring systems, interactive educational tools, and mental health diagnostics. In these real-world settings, the wide variability in factors such as illumination, camera quality, pose, and occlusions induces significant domain shifts between the training and testing distributions. As a consequence, models trained on well-curated datasets often fail to generalize to the diverse conditions encountered in practical deployments, necessitating robust adaptation strategies that can operate under such challenging circumstances.

Domain shift remains a formidable challenge as the statistical properties of test data frequently deviate from those observed during training, leading to considerable performance degradation in deep learning models. Traditional domain adaptation and generalization techniques often rely on access to target domain data or extensive retraining, which may be impractical in dynamic or privacy-sensitive environments. Test-time adaptation (TTA) has emerged as a promising alternative that allows a pre-trained model to refine its parameters at inference using unlabeled test samples. This approach mitigates the need for large-scale re-annotation and re-training, enabling the model to continuously adjust to evolving conditions during deployment.

A promising direction within TTA leverages the temporal consistency inherent in video data. Methods that enforce temporal coherence assume that, in high frame-rate videos, consecutive frames exhibit gradual changes in appearance and, consequently, in model predictions \cite{mutlu2023tempt}. Such temporal consistency provides a self-supervision signal that can be exploited to adapt model parameters at test time. However, prior work in this domain typically relies on manually selecting a fixed subset of parameters. This manual selection, while effective in certain scenarios, may be computationally expensive and suboptimal, as it does not account for the dynamic importance of different parameters under varying test conditions.

In contrast, the sensitivity of a network’s loss with respect to its parameters, as characterized by the Fisher Information Matrix \cite{kirkpatrick2017overcoming}, offers a principled metric for parameter importance. Fisher-based measures have been successfully employed in continual learning to mitigate catastrophic forgetting and have recently been explored in the context of test-time adaptation \cite{niu2022efficient}. By quantifying the contribution of each weight to the overall loss, Fisher scores enable a data-driven, selective adaptation strategy.

This paper introduces a novel Fisher-driven selective adaptation framework for TTA. The proposed method integrates Fisher information into the temporal consistency paradigm. By leveraging Fisher scores to dynamically assess parameter importance, our approach selectively updates only those weights that are most critical to model performance under test conditions. This principled, data-driven selection mechanism not only streamlines the adaptation process but also addresses the rigidity and inefficiency of manual layer selection. 

The proposed Fisher-driven selective adaptation framework offers several key advantages. First, by focusing on dynamically selected, high-importance weights, the method significantly reduces computational overhead compared to approaches that update a large, fixed subset of parameters. Second, the selective updating process enhances model robustness, as evidenced by improved F1 scores in facial expression recognition tasks under diverse domain shifts. Finally, the approach is inherently well-suited for in-the-wild affective behavior analysis, providing a scalable and efficient solution for real-world applications. Experimental results on the AffWild2 benchmark dataset \cite{kollias2019deep,kollias2019expression,kollias2019face,kollias2020analysing,kollias2021affect,kollias2021analysing,kollias2021distribution,kollias2022abaw,kollias2023abaw,kollias2023abaw2,kollias2023multi,kollias20246th,kollias20247th,kollias2024distribution,zafeiriou2017aff}, demonstrate that the proposed approach yields competitive performance improvements over both the base supervised model and the base TTA method.

\section{Related Work}
\label{sec:relatedwork}

%-------------------------------------------------------------------------
\subsection{Facial Expression Recognition}
Facial expression recognition (FER) is a central problem in affective computing, with applications ranging from user engagement in interfaces and gaming to driver monitoring, mental health assessment, and security systems. In real-world settings, FER models must cope with large intra-class variations caused by changes in lighting, head pose, occlusions (e.g. glasses, hands), and individual differences. Faces can appear in varied head poses and under diverse illumination, often partially occluded by objects or self-blocking (e.g. hand on face). Subject-specific differences (identity, age, gender) further modulate facial appearance, meaning the same expression can look different on different people, while different expressions can sometimes look similar. Such intra-class variance and inter-class ambiguity demand FER models that generalize across pose, lighting, occlusion, and identity variations. Handling these factors is critical for FER systems to perform reliably in unconstrained settings.

Early deep learning approaches for video-based FER predominantly employed convolutional neural networks (CNNs) to learn robust representations in unconstrained environments \cite{8453893,8576656,9075283,qian2023computervisionestimationemotion}. More recently, transformer-based architectures have also been leveraged to capture long-range dependencies and relational features among facial regions, further enhancing recognition accuracy \cite{xue2021transfer}. Minimal work has addressed test-time adaptation (TTA) for FER, despite its promise in adapting to dynamic conditions such as lighting, pose, and occlusion without extra labeled data \cite{mutlu2023tempt}. In this work, we extend the temporal-consistency-based TTA paradigm to develop a more robust and adaptive FER system for real-world, unconstrained environments.

\subsection{Test-Time Adaptation}
Test-time adaptation (TTA) methods adjust a model during inference using only unlabeled test data, thereby addressing domain shifts that static training cannot anticipate. Early approaches in this space updated the running statistics of batch normalization (BN) layers \cite{bn1, bn_adapt} to reflect the new test data distribution. Later, methods employed auxiliary self-supervised tasks to adapt backbone parameters \cite{ttt}, while others refined this strategy by minimizing output entropy and selectively updating BN weights—capitalizing on their high expressive power \cite{tent, bn_expressive}. More recent techniques have also explored meta-learning solutions \cite{arm} and combined image augmentation with entropy minimization to handle larger domain shifts \cite{memo, shot}. 

While effective on static data, these approaches generally overlook the rich temporal correlations inherent in video data, which are crucial for tasks like FER. Only a few works, such as \cite{tent}, have ventured into continual adaptation for online streams, with \cite{cotta} proposing augmentation consistency and \cite{note} introducing a novel normalization layer to address non-independent and non-identically distributed (non-i.i.d.) test conditions.

Temporal-consistency-based approaches have demonstrated that leveraging smooth transitions in sequential frames can yield more stable predictions \cite{mutlu2023tempt}. However, existing temporal-based methods often rely on manual or naïve selection of the subset of parameters to update, lacking a principled strategy to determine which weights are most beneficial to adapt.

In this work, we extend the temporal-consistency paradigm by introducing dynamic selection of model parameters. This approach eliminates the need for manual weight selection and avoids blindly choosing a subset, leading to computationally more efficient and better-adapted updates specifically tailored for video data.

\begin{table*}[h]
\centering
\begin{tabular}{|l|r|r|r|r|r|r|r|r|r|}
\hline
 & Happiness & Surprise & Neutral & Disgust & Anger & Fear & Sadness & Other & Total \\
\hline
Affwild2 & 2622 & 5540 & 44676 & 7851 & 32962 & 9730 & 3296 & 31412 & 138089 \\
Affectnet & 5670 & 19325 & 55670 & 19650 & 118605 & 11647 & 3626 & 0 & 234193 \\
RAF-DB & 347 & 846 & 3096 & 2390 & 5771 & 1571 & 865 & 0 & 14886 \\
\hline
\textbf{Total} & \textbf{8639} & \textbf{25711} & \textbf{103442} & \textbf{29891} & \textbf{157338} & \textbf{22948} & \textbf{7787} & \textbf{31412} & \textbf{387168} \\
\hline
\end{tabular}
\caption{Distribution of facial emotion recognition datasets labels}
\label{tab:data}
\end{table*} 

\subsection{Fisher Information Matrix}
The Fisher Information Matrix (FIM) quantifies the sensitivity of a model’s loss to changes in each parameter, making it an effective measure of weight importance. This concept is rooted in Bayesian statistics and was initially popularized through the use of Laplace approximations in neural network learning, as discussed by MacKay \cite{mackay1992practical}. Under this approximation, the posterior distribution of model parameters is approximated by a Gaussian whose precision (inverse variance) is given by the FIM’s diagonal elements. As noted by Pascanu and Bengio \cite{pascanu2013revisiting}, the diagonal elements of the FIM effectively capture the curvature of the loss surface near a minimum. A higher Fisher value for a weight implies that the model’s loss is highly sensitive to changes in that parameter, suggesting that it is significantly contributing to the learned function.

This idea of parameter importance estimation via Fisher information has been widely applied to preserve important knowledge in various learning paradigms. In the realm of continual learning, Kirkpatrick et al. \cite{kirkpatrick2017overcoming} introduced Elastic Weight Consolidation (EWC), which uses the Fisher information of each parameter to regularize learning on new tasks. Further extended by works such as \cite{li2017learning}, \cite{rolnick2019experience}, \cite{farajtabar2020orthogonal}, \cite{niu2021disturbance}, \cite{mittal2021essentials}, and \cite{xu2020forget}, Fisher-based regularization is employed to prevent catastrophic forgetting by constraining critical weights from deviating too far from their previously learned values. 

More recently, the concept has has been adopted in test-time adaptation frameworks. Niu et al.\cite{niu2022efficient} proposed an efficient test-time adaptation method that explicitly addresses forgetting by computing Fisher information on-the-fly for the test distribution. In a similar vein, Brahma et al. \cite{brahma2023probabilistic} leverage Fisher information in a lifelong test-time adaptation framework. They stochastically update the model on a stream of target data and use a Fisher-based criterion to decide when to reset certain parameters back to their source values, rather than resetting weights at random. 

These works demonstrate that Fisher information is a powerful tool for selective learning under domain shift. Motivated by this, our approach applies a Fisher-based selection strategy within a test-time adaptation framework, enabling the model to adapt to new facial expression domains while updating only the parameters most vital to its base performance. To the best of our knowledge, there has been no prior work on selective test-time adaptation driven by Fisher scores. Existing TTA solutions either treat all parameters equally or rely on manual selection to decide which weights to update, often leading to suboptimal or computationally heavy outcomes. Our proposed method addresses this gap by systematically computing Fisher scores for each parameter at test time, then selectively updating only the most critical parameters. This strategy not only minimizes computational overhead but also robustly updates the core knowledge necessary for accurate facial expression recognition under domain shifts, representing a novel contribution to the field.

\section{Method}
\label{sec:methods}

\subsection{Datasets and Preprocessing}
To build a comprehensive and diverse training set, several well-known datasets from the facial expression recognition literature were merged, including Affwild2 \cite{kollias2019deep,kollias2019expression,kollias2019face,kollias2020analysing,kollias2021affect,kollias2021analysing,kollias2021distribution,kollias2022abaw,kollias2023abaw,kollias2023abaw2,kollias2023multi,kollias20246th,kollias20247th,kollias2024distribution,zafeiriou2017aff}, Affectnet \cite{mollahosseini2017affectnet}, and the Real-world Affective Faces Database (RAF-DB) \cite{rafdb1, rafdb2}. The model is designed to classify images into seven basic emotions—commonly referred to as Ekman emotions \cite{ekman}—along with an additional “other” category for expressions that do not fall into these predefined classes.
Affwild2 is notably larger than the other datasets and exhibits a significant imbalance in label distribution. To address this issue, a random sampling strategy is applied, limiting the number of frames per video per emotion category to 300, following the approach in \cite{mutlu2023tempt}. The specific label distribution after this sampling is detailed in Table~\ref{tab:data}.
Since Affwild2 already provides cropped and aligned images, and the remaining datasets are available only as cropped images, no further spatial preprocessing is necessary. All images are then resized to $112\text{px}\times112\text{px}$ using antialiasing. Additionally, we implement common image augmentation techniques during training, such as random horizontal flip, color jitter, channel dropout, brightness and contrast shift, blur, and histogram equalization.

\begin{figure*}[t]
  \centering
  \includegraphics[width=\textwidth]{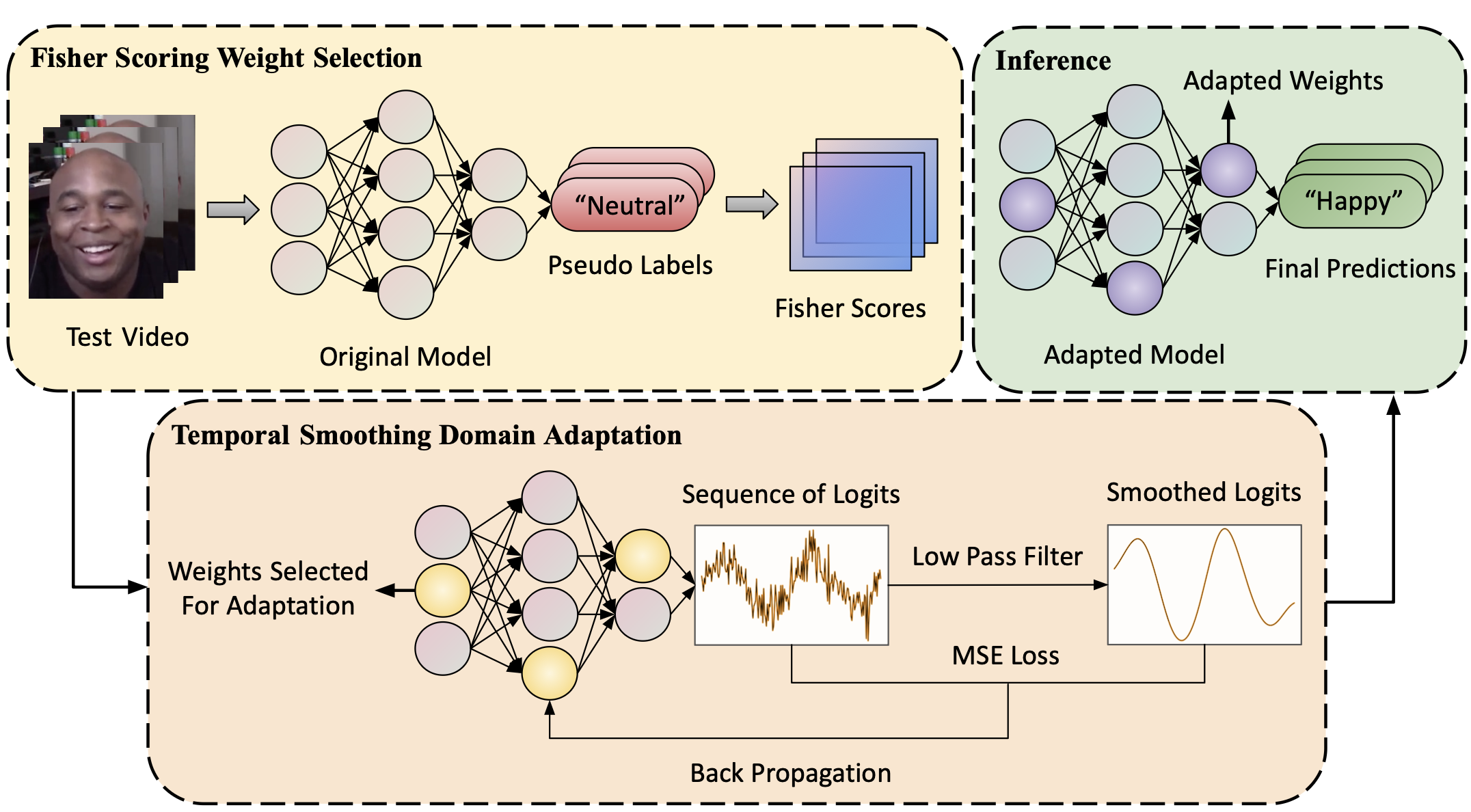}
  \caption{Overview of our proposed selective test-time domain adaptation framework using Fisher information. The method consists of three key components: 1) Fisher Scoring Weight Selection (top left), which processes sample frames from a test video through the original model to generate pseudo-labels and compute Fisher importance scores for model parameters; 2) Temporal Smoothing Domain Adaptation (bottom), which selectively updates only the most important weights (highlighted in yellow) by minimizing the difference between original model logits and their temporally smoothed versions using a low-pass filter; and 3) Inference (top right), where the adapted model with updated weights produces final expression predictions.}
  \label{fig:methods}
\end{figure*}

\subsection{Model}
The model architecture and training strategy follow prior work on test-time adaptation for facial expression recognition \cite{mutlu2023tempt}. Predictions are made on individual video frames, allowing the use of well-established image-processing architectures. Given their reliable performance and stable training, models from the ResNet \cite{resnet} family are employed, incorporating enhancements such as aggregated residual transformations \cite{resnext} and squeeze-and-excitation blocks \cite{hu2018squeeze}. The resulting embeddings are then passed through two fully connected layers, with the final (output) layer undergoing both weight and input normalization \cite{salimans2016weight} to mitigate overconfidence, ensure smoothness, and enhance generalization.

A critical challenge in this setting is the significant class imbalance, which must be addressed for effective supervised learning. Traditional solutions such as label weighting, up-sampling, or down-sampling often prove inadequate in certain scenarios. To address this, Label-Distribution-Aware Margin Loss (LDAM) \cite{ldam} is adopted. Unlike standard sample weighting, which scales the loss multiplicatively, LDAM adjusts the class margins based on the frequency of each class. The precise formulation is given in \cref{ldam_loss}, where $z$ is the unnormalized prediction vector, $y$ is the ground truth label, $n_j$ denotes the number of samples in class $j$, and $C$ is a temperature-like hyperparameter that modulates the margin's impact. By enforcing larger margins for minority classes, LDAM enhances the model's robustness and reduces overfitting. 

\begin{equation}
\begin{split}
\label{ldam_loss}
    \mathcal{L}(z,y) &= - \log\frac{e^{z_y-\Delta_y}}{e^{z_y-\Delta_y} + \sum_{j\neq y}e^{z_j}} \\
    \text{where  } \Delta_j &= \frac{C}{n_j^{1/4}} \text{ for } j\in \{1, \dots, k \}
\end{split}
\end{equation}

Supervised training is then conducted using back-propagation with the defined LDAM loss to account for the skewed label distribution. Optimization is performed using the Adam \cite{adam} optimizer with weight decay \cite{adamw}, following a step-decay schedule for the learning rate. The modeling and training processes are implemented in the PyTorch \cite{paszke2019pytorch} framework on NVIDIA V100 GPUs.

\subsection{Temporal Smoothing Domain Adaptation}
\label{subsec:tempt}

Following the methodology in \cite{mutlu2023tempt}, we adopt a temporal-consistency-based domain adaptation strategy to enhance prediction stability across video frames. Since 2D CNNs are trained on individual static images rather than video sequences, they lack an inherent bias for smooth and consistent predictions across consecutive frames. It has been observed that these models tend to produce outputs with pronounced high-frequency components. However, when a low-pass filter is applied to the predictions, the resulting outputs appear more coherent. This observation is leveraged by using the filtered predictions as a supervision signal to tune the network, thereby encouraging temporal consistency. Specifically, the model's outputs are first smoothed over time using a low-pass filter. This smoothed version is designated as the target, and the mean-squared error between the original and filtered predictions is computed. This error signal is then backpropagated to update a selected subset of model parameters.

Let $x^{(t)}\in \mathbb{R}^{112\times112\times3}$ denote the $t^{th}$ frame of a video, and let $f(.):\mathbb{R}^{112\times112\times3}\rightarrow\mathbb{R}^{8}$ represent the neural network. It is hypothesized that ensuring coherence between predictions on consecutive frames can serve as an implicit Jacobian regularizer. Previous work \cite{hoffman2019robust} has shown that regularizing the Frobenius norm of the network's input-output Jacobian can help achieve flatter minima and improve robustness to input variations. When the video’s frame rate is sufficiently high, the Jacobian can be approximated as follows:

\begin{equation}
\label{jacobian}
    J_{i,j}(x^{(t)}) = \frac{\partial f_i(x^{(t)})}{\partial x_j^{(t)}} \approx \frac{f_i(x^{(t)})-f_i(x^{(t-1)})}{x_j^{(t)}-x_j^{(t-1)}}
\end{equation}

Thus, minimizing the Frobenius norm of the Jacobian essentially amounts to reducing the difference between consecutive frame predictions:

\begin{equation}
\begin{split}
\label{frob_minimize}
    \min\|J(x^{(t)})\|_F &\equiv \min\sum_{i,j} J^2_{i,j}(x^{(t)}) \\ &\equiv \min \|f_i(x^{(t)})-f_i(x^{(t-1)})\|
\end{split}    
\end{equation}

Empirical analysis reveals that the distribution of prediction differences is heavy-tailed. This behavior is mainly due to abrupt changes in predictions caused by issues like improper cropping or sudden activation shifts stemming from model imperfections. Directly using the formulation in \cref{frob_minimize} can lead to instability during training because these outliers disrupted the optimization process. To mitigate this, an equivalent formulation is adopted: all frames are first processed to obtain a set of unnormalized scores $y^{(t)}\in\mathbb{R}^{8}$ and then a self-supervision loss function is defined as follows:

\begin{equation}
\label{eq:loss_signal}
    \mathcal{L}(y) = \sum_{t} \| y^{(t)} - LPF(y)^{(t)} \|
\end{equation}

Here, $LPF(.)$ represents a low-pass filter (a median filter is used in experiments for its robustness to outliers). For long videos, processing every frame for adaptation can be computationally expensive. To address this, a sliding window is employed to count changes in the model’s predictions, selecting only those regions exhibiting the most significant variations to form the training batch. The loss function is then updated as in \cref{eq:loss_signal_batch}, where $\mathcal{R}$ denotes the set of selected regions and $r$ represents a range of frames within a region:

\begin{equation}
\label{eq:loss_signal_batch}
    \mathcal{L}(y) = \sum_{r \in \mathcal{R}} \sum_{t \in r} \| y^{(t)} - LPF(y)^{(t)} \|
\end{equation}

Because this loss is differentiable, backpropagation can be used to update model parameters. The selection of which parameters to update is critical, as it impacts the model's expressivity and the effectiveness of the adaptation. In this work, we investigate whether a more systematic and dynamic approach to parameter selection exists, instead of relying on manual choices. To achieve this, we employ a Fisher scoring weight selection strategy, which is discussed in depth in Section~\ref{subsec:fisher}. Finally, the adaptation process is carried out using the AdamW optimizer with a learning rate of 0.0001, and we perform 4 gradient steps—a setting that we found to be empirically optimal.

\begin{table*}[ht]
\centering
\setlength{\tabcolsep}{5pt}
\renewcommand{\arraystretch}{1.2}
\small
\begin{tabular}{l | c | c | c c c c | c c}
\toprule
& & \textbf{TENT \cite{tent}} & \multicolumn{4}{c|}{\textbf{Temporal Smoothing Adaptation}} & \multicolumn{2}{c}{\textbf{Fisher Scoring Weight Selection}} \\
\cmidrule(lr){3-3} \cmidrule(lr){4-7} \cmidrule(lr){8-9}
& \textbf{\shortstack{Base Model\\(SE-ResNeXt-101\\Backbone)}} & \textbf{\shortstack{BatchNorm\\Layers}} & \textbf{\shortstack{All\\Layers}} & \textbf{\shortstack{Early\\Layers}} & \textbf{\shortstack{Mid\\Layers}} & \textbf{\shortstack{Late\\Layers}} & \textbf{\shortstack{0.2\% of\\All\\Weights}} & \textbf{\shortstack{5\% of\\Early Layers\\Weights}} \\
\midrule
\textbf{F1 Score} & 0.325 & 0.269 & 0.300 & 0.345 & 0.288 & 0.330 & 0.341 & 0.350 \\
& & \textcolor{red}{(–17.2\%)} & \textcolor{red}{(–7.7\%)} & \textcolor{green}{(+6.2\%)} & \textcolor{red}{(–11.4\%)} & \textcolor{green}{(+1.54\%)} & \textcolor{green}{(+4.9\%)} & \textcolor{green}{(+7.7\%)} \\
\midrule
\textbf{Weights Adapted} & 0 & \textcolor{green}{203k} & \textcolor{red}{91.8M} & \textcolor{green}{456k} & \textcolor{red}{91.3M} & \textcolor{red}{30.6M} & \textcolor{green}{183k} & \textcolor{green}{22k} \\
% \textbf{Adapted} & (Supervised Method) & & & & & & \\
\bottomrule
\end{tabular}
\caption{Performance comparison of different test-time adaptation methods on facial expression recognition. The table shows F1 scores and number of weights adapted for each method, with percentage changes relative to the base model in parentheses. Our Fisher-based selective approaches (rightmost columns) achieve the best performance while adapting significantly fewer parameters.}
\label{tab:performance}
\end{table*}

\subsection{Fisher Scoring Weight Selection}
\label{subsec:fisher}

In this section, we propose a novel Fisher scoring-based strategy to identify and select the most important model weights, updating only these during backpropagation. Let $\phi(\theta_i)$ represent the importance of the parameter $\theta_i$ , which we estimate using the diagonal Fisher information matrix as done in elastic weight consolidation \cite{kirkpatrick2017overcoming}. Typically, computing the Fisher information requires a set of labeled in-distribution (ID) training samples. However, in our setting we do not have access to the training data, and our test samples are unlabeled, making it challenging to assess weight importance.

Inspired by the method introduced in \cite{niu2022efficient}, we address this challenge by first collecting a small set of \(N\) unlabeled sample frames \(x^{(t)}\) from the test video. We then use the originally trained model \(f_{\Theta}(\cdot)\) to predict these samples, thereby obtaining pseudo-labels \(\hat{y}^{(t)}\). With these, we construct a pseudo-labeled set of sampled frames \(\mathcal{Q}\). We calculate the Fisher importance of each model weight \(\theta_i\) as an average of its Fisher scores over this set using the following formula:

\begin{equation}
\label{eq:fisher}
\phi(\theta_i) = \frac{1}{N} \sum_{x^{(t)} \in \mathcal{Q}} \left( \frac{\partial}{\partial \theta_i} \mathcal{L}_{\text{CE}}(f_{\Theta}(x^{(t)}), \hat{y}^{(t)}) \right)^2
\end{equation} where \(\mathcal{L}_{\text{CE}}\) denotes the cross-entropy loss. 

We experimented with different numbers of frames sampled \(N\) (see Section~\ref{sec:experiments} for detailed results), but in most cases a 1-3 frames per video were sufficient.

After calculating the Fisher scores as specified in Eq.~\ref{eq:fisher}, we rank the model’s weights by importance and select the top $k\%$ of weights for updating during backpropagation. In this work, we experimented with adapting different percentages of model weights, both from early layers and across all layers, with detailed results provided in Section~\ref{sec:experiments}. The Fisher-driven selected weights then undergo the temporal smoothing domain adaptation module, as described in Section~\ref{subsec:tempt}, and the adapted model is subsequently used for label prediction. A comprehensive overview of our complete methodology and framework is presented in Fig.~\ref{fig:methods}.

\section{Experiments}
\label{sec:experiments}

We evaluate our Fisher-driven selective adaptation framework on the challenging task of facial expression recognition (FER) in-the-wild using the AffWild2 \cite{kollias2019deep,kollias2019expression,kollias2019face,kollias2020analysing,kollias2021affect,kollias2021analysing,kollias2021distribution,kollias2022abaw,kollias2023abaw,kollias2023abaw2,kollias2023multi,kollias20246th,kollias20247th,kollias2024distribution,zafeiriou2017aff} validation dataset. In our experiments, we report the macro F1 score calculated across predictions on all frames of all videos, which is a standard evaluation metric for expression recognition tasks, especially in cases with class imbalance.

\subsection{Experimental Setup}

Our base model employs an SE-ResNeXt-101 \cite{hu2018squeeze} backbone trained on a combination of facial expression recognition datasets (AffWild2 \cite{kollias2019deep,kollias2019expression,kollias2019face,kollias2020analysing,kollias2021affect,kollias2021analysing,kollias2021distribution,kollias2022abaw,kollias2023abaw,kollias2023abaw2,kollias2023multi,kollias20246th,kollias20247th,kollias2024distribution,zafeiriou2017aff}, AffectNet \cite{mollahosseini2017affectnet}, and RAF-DB\cite{rafdb1,rafdb2}). This approach provides a strong foundation model that has already learned to recognize expressions across a diverse range of subjects and conditions, serving as a challenging baseline to improve upon.

In our experiments, we compare several test-time adaptation methods. First, we evaluate TENT \cite{tent}, which leverages entropy minimization and has shown competitive performance on various benchmarks. We also assess Temporal Smoothing Adaptation, derived from the TempT framework \cite{mutlu2023tempt}. For this approach, we explore multiple weight selection strategies in addition to our newly proposed Fisher Scoring Weight Selection method.

\subsection{Performance Comparison}

Table~\ref{tab:performance} summarizes the comparative performance of all methods, including the F1 scores achieved on the validation set and the number of weights adapted during the test-time adaptation process.
The results reveal several key findings. First, we observe that TENT \cite{tent}, despite its success in other test-time adaptation scenarios, significantly degrades performance on facial expression recognition, with a substantial 17.2\% decrease in F1 score compared to the base model. This highlights the challenges of applying general-purpose adaptation methods to the emotionally nuanced and temporally sensitive domain of facial expressions.

For the regular Temporal Smoothing Adaptation approach, we find that performance is heavily dependent on the selection of which subset of model layers and weights to adapt. In our experiments, we manually selected different subsets for adaptation:

\begin{itemize}
    \item \textbf{Early layers:} Including up to the end of the first layer of the SE-ResNeXt model
    \item \textbf{Mid layers:} Including the 2nd and 3rd layers
    \item \textbf{Late layers:} Including the 4th layer until the end of the model
    \item \textbf{All layers:} The entire model
\end{itemize}

The results demonstrate the critical impact of layer selection. While adapting early layers yields a 6.2\% improvement over the base model, adapting mid layers causes an 11.4\% performance drop. Adapting late layers provides a modest 1.54\% improvement, and adapting all layers actually degrades performance by 7.7\%. This inconsistency underscores a significant limitation of the regular adaptation approach: it requires extensive experimentation to identify which layers to adapt for optimal performance, making it impractical for real-world deployment.
Furthermore, most variations of regular temporal smoothing adaptation require adapting a substantial number of weights—in the order of millions—with the exception of the early layers version, which still necessitates updating 456,000 parameters. This high computational demand presents a significant barrier for deployment in resource-constrained environments or real-time applications.

\begin{figure*}[t]
  \centering
  \includegraphics[width=0.49\textwidth]{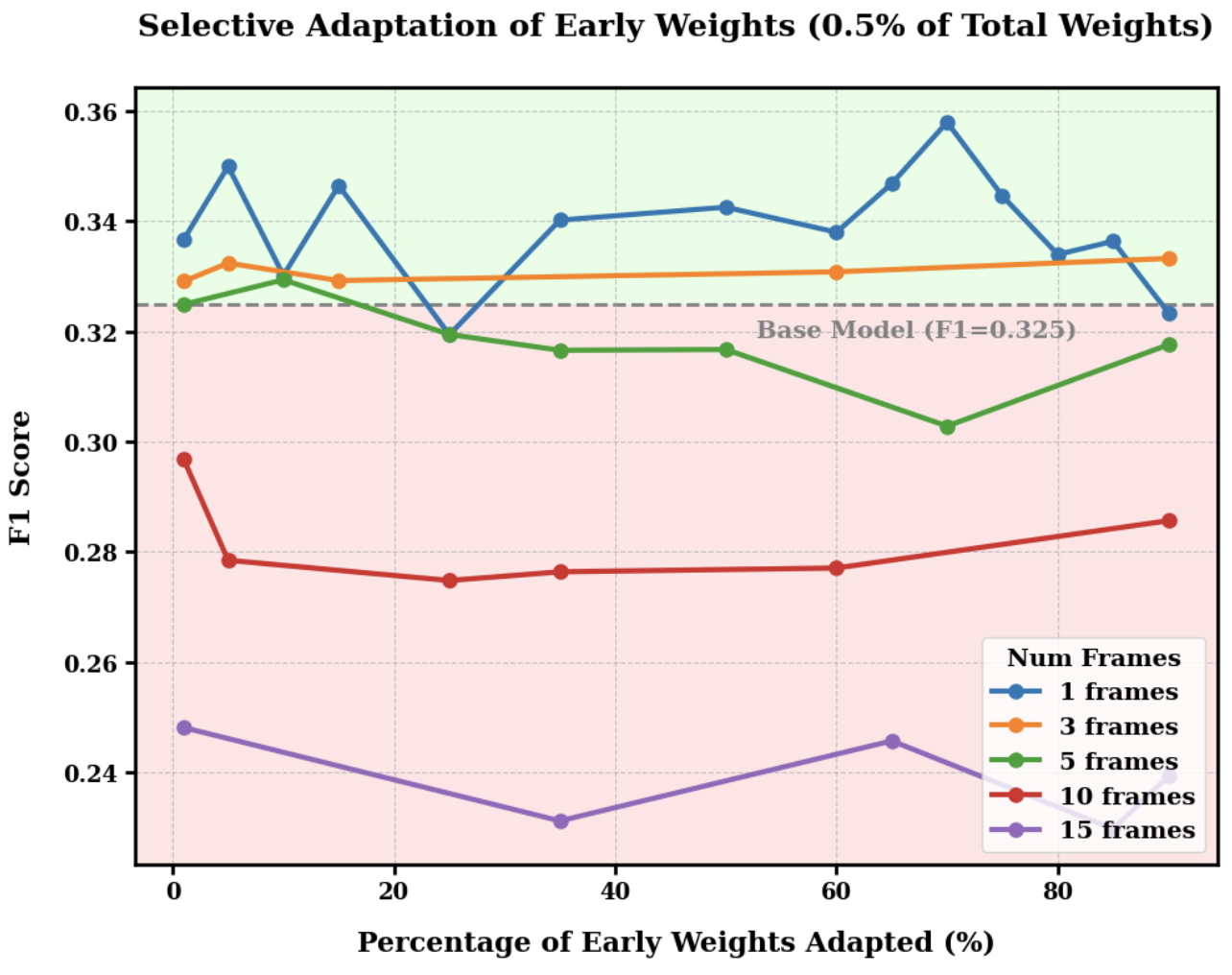}
  \hfill
  \includegraphics[width=0.49\textwidth]{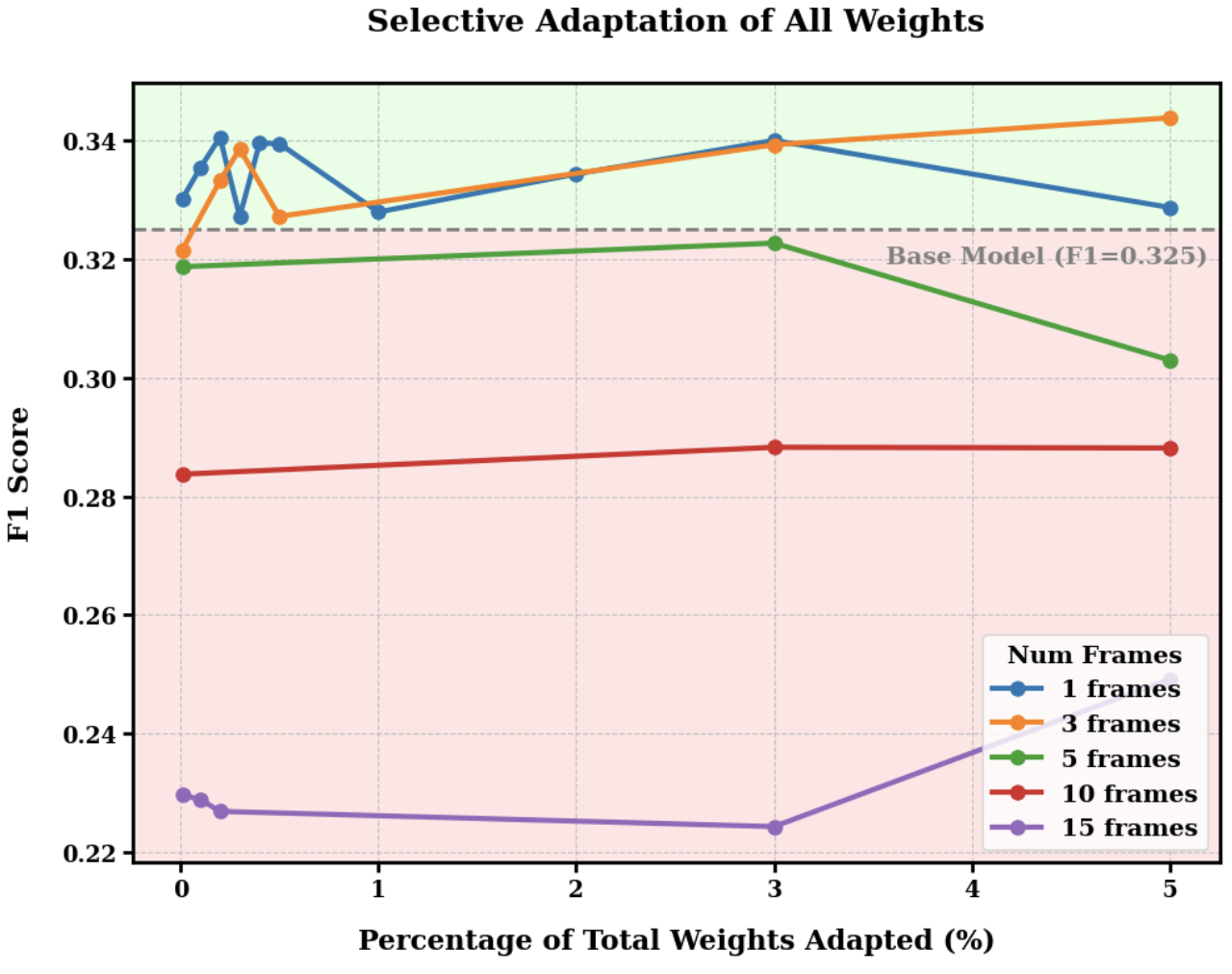}
  \caption{Ablation study on the effect of percentage threshold and frame sampling on adaptation performance. Left: Results when selecting from early layer weights (0.5\% of total model weights). Right: Results when selecting from all model weights. The horizontal dashed line indicates base model performance (F1=0.325), with the green region above showing improvement and the red region below showing degradation.}
  \label{fig:ablation_fisher}
\end{figure*}

\subsection{Fisher-Based Selective Adaptation}

Our proposed Fisher scoring weight selection method addresses both the performance inconsistency and computational inefficiency of previous approaches. The method selects the most important weights to adapt using Fisher scores based on pseudo labels of the model on the target domain at test time. This approach personalizes the weight selection process to each individual video at test time, making the adaptation both more effective and computationally efficient.
We implemented our Fisher-based selection using two strategies:

\begin{itemize}
    \item Selecting the top 0.2\% of weights across the entire model based on Fisher scores
    \item Selecting the top 5\% of weights from the early layers of the model
\end{itemize}

Both strategies yield impressive results. When selecting from the entire model, our method achieves an F1 score of 0.341, representing a 4.9\% improvement over the base model. When focusing on early layers, it reaches an F1 score of 0.350, a 7.7\% improvement that surpasses all other adaptation approaches. This performance is particularly notable considering that it outperforms the base SE-ResNeXt-101 model, which is already a very large and powerful network for facial expression recognition.

Perhaps most significantly, our Fisher-based approach dramatically reduces the computational burden of adaptation. When selecting from early layers, it requires updating only 22,000 weights out of the model's total 92,000,000 parameters—more than 20 times fewer than the regular early layers adaptation and over 4,000 times fewer than adapting the entire model. Even when selecting from all layers, our method still maintains a minimal footprint, adapting just 183,000 weights.

These results demonstrate that our Fisher-driven selective adaptation framework not only improves recognition performance but does so with substantially greater efficiency than existing approaches. By dynamically identifying the most important weights for each test video, the method eliminates the need for manual layer selection while achieving superior performance with minimal computational overhead.

\subsection{Ablation Studies}

To further investigate the effectiveness of our Fisher-driven selective adaptation approach, we conducted ablation studies examining two critical aspects of the method: (1) the impact of the percentage threshold used for weight selection, and (2) the effect of averaging Fisher scores across multiple frames when estimating parameter importance.

\subsubsection{Impact of Percentage Threshold}

Fig.~\ref{fig:ablation_fisher} presents the performance trends of our approach across different percentage thresholds for both adaptation strategies: selecting from early layers only (left plot) and selecting from all model weights (right plot). The horizontal dashed line represents the base model performance (F1 = 0.325), with the green region above indicating improvement and the red region below showing performance degradation.

We observe that performance remains robust across a wide range of percentage thresholds when using a small number of frames (1-3) for Fisher score calculation. This relative insensitivity to the exact percentage threshold suggests that a very limited subset of weights is responsible for the majority of the adaptation benefits. This finding strongly supports our core hypothesis that selective adaptation of the most critical parameters can yield substantial performance improvements while minimizing computational overhead.

\subsubsection{Effect of Frame Sampling}

Perhaps counterintuitively, our results demonstrate that averaging Fisher scores over multiple frames does not provide additional benefits and frequently leads to performance degradation. As shown in Fig.~\ref{fig:ablation_fisher}, models using Fisher scores calculated from 5, 10, or 15 frames consistently underperform compared to those using only 1 or 3 frames, with many configurations falling below baseline performance.

This phenomenon can be explained by considering the nature of Fisher information in the context of facial expression recognition. Fisher scores measure the sensitivity of the loss with respect to parameters for specific inputs. When averaging across multiple frames, especially in videos with expression transitions or significant variations, the resulting averaged Fisher scores may no longer accurately reflect the most critical parameters for any particular frame. This "dilution effect" appears to diminish the precision of the parameter importance estimation, leading to suboptimal adaptation.

In contrast, using Fisher scores from a single frame or a very small number of frames (1-3) maintains the specificity of the importance estimation. The slight performance fluctuations observed with these lower frame counts across different percentage thresholds likely reflect the sensitivity of Fisher scores to the specific content of the sampled frames. Despite these fluctuations, the performance remains consistently above the baseline, highlighting the robustness of the overall approach.

The superior performance of low-frame sampling validates our selective adaptation strategy while suggesting that parameter importance can be effectively estimated from minimal data. This has significant practical implications, as it enables extremely efficient adaptation with minimal computational overhead—sampling just 1-3 frames and updating less than 0.5\% of the model's parameters is sufficient to achieve substantial performance improvements.

\section{Conclusion}

In this paper, we introduced a novel Fisher-driven selective adaptation framework for test-time domain adaptation in facial expression recognition. Our approach addresses critical limitations of existing test-time adaptation methods by employing Fisher information to dynamically identify and update only the most important model parameters during adaptation.

Our experimental results on the challenging AffWild2 benchmark demonstrate that the proposed method consistently outperforms both the baseline model and existing adaptation techniques. This performance gain is particularly significant considering that it surpasses methods that adapt substantially more parameters, including those that update the entire network.

The computational efficiency of our approach represents a major advancement for practical applications. By reducing the number of adapted parameters from millions to a few thousand, our method enables efficient adaptation even in resource-constrained environments. This dramatic reduction in computational overhead, combined with improved performance, makes our approach particularly well-suited for real-world deployment.
% in applications such as driver monitoring systems, mental health diagnostics, and educational tools where latency and power consumption are critical considerations.

Our ablation studies revealed several important insights. Most notably, we found that averaging Fisher scores across multiple frames often degrades performance compared to using scores from just a single frame or very few frames. This finding suggests that parameter importance estimation is most effective when focused on specific instances rather than averaged across diverse inputs—particularly in domains like facial expression recognition where subtle temporal variations carry significant semantic meaning.

Beyond facial expression recognition, the proposed selective adaptation framework offers a principled approach to test-time adaptation that could be extended to other computer vision tasks facing domain shift challenges. By eliminating the need for manual layer selection and providing a data-driven mechanism for identifying adaptation-critical parameters, our method addresses the fundamental limitations of existing approaches that rely on heuristics or exhaustive experimentation.

While our work demonstrates significant improvements, several avenues for future research remain open. Investigating dynamic adjustment of the percentage threshold based on domain characteristics, exploring alternative parameter importance metrics beyond Fisher information, and extending the approach to more diverse visual understanding tasks could further enhance the method's applicability. Additionally, investigating how selective adaptation might be combined with other complementary strategies such as self-supervision or entropy minimization represents a promising direction for future exploration.

In conclusion, our Fisher-driven selective adaptation framework represents a significant step toward more efficient, effective, and practical test-time adaptation for facial expression recognition in unconstrained environments, opening new possibilities for robust affective computing in real-world applications.

\bibliographystyle{ieeenat_fullname}
% \bibliography{main}


\end{document}